\pdfoutput=1

\documentclass[11pt]{article}

\usepackage{color}
\definecolor{wildstrawberry}{rgb}{1.0, 0.26, 0.64}

\newcommand{\dataname}{ExPUNations\xspace}

\usepackage{times}
\usepackage{subcaption}
\usepackage{latexsym}
\usepackage{xspace}
\usepackage{booktabs}
\usepackage{multirow}
\usepackage{fontawesome5}
\usepackage{graphicx}
\usepackage{tikz}
\usetikzlibrary{plotmarks}
\usepackage{fontawesome5}
\usepackage{quoting}

\usepackage{algorithm}
\usepackage{algorithmic}
\usepackage{soul}
\usepackage{enumitem}
\usepackage{todonotes}

\newcommand{\draftonly}[1]{#1}
\renewcommand{\draftonly}[1]{}

\newcommand{\RNum}[1]{\uppercase\expandafter{\romannumeral #1\relax}}

\usepackage{emnlp2022}

\usepackage{times}
\usepackage{latexsym}

\usepackage[T1]{fontenc}

\usepackage[utf8]{inputenc}

\usepackage{microtype}

\title{\dataname: Augmenting Puns with Keywords and Explanations}

\author{Jiao Sun$^1$\thanks{\xspace\xspace Work done during Jiao's internship at Amazon.}\protect\phantom{\footnotesize 1}\textsuperscript{}\thanks{\xspace\xspace Both authors equally contributed to the paper.} \, Anjali Narayan-Chen$^{2}$\footnotemark[2] \, Shereen Oraby$^{2}$ \, Alessandra Cervone$^{2}$  \,\\  \textbf{Tagyoung Chung}$^{2}$ \, \textbf{Jing Huang}$^{2}$ \, \textbf{Yang Liu}$^{2}$ \, \textbf{Nanyun Peng}$^{2,3}$ \\
$^1$University of Southern California \\
$^2$Amazon Alexa AI\\
$^3$University of California, Los Angeles \\
\texttt{jiaosun@usc.edu} \\
\texttt{\{naraanja,orabys,cervon,tagyoung,jhuangz,yangliud\}@amazon.com }\\
\texttt{violetpeng@cs.ucla.edu}
}

\begin{document}
\maketitle
\begin{abstract}
The tasks of humor understanding and generation are challenging and subjective even for humans, requiring commonsense and real-world knowledge to master. Puns, in particular, add the challenge of fusing that knowledge with the ability to interpret lexical-semantic ambiguity. In this paper, we present the ExPUNations (ExPUN) dataset, in which we augment an existing dataset of puns with detailed crowdsourced annotations of keywords denoting the most distinctive words that make the text funny, pun explanations describing why the text is funny, and fine-grained funniness ratings. This is the first humor dataset with such extensive and fine-grained annotations specifically for puns. Based on these annotations, we propose two tasks: explanation generation to aid with pun classification and keyword-conditioned pun generation, to challenge the current state-of-the-art natural language understanding and generation models' ability to understand and generate humor. We showcase that the annotated keywords we collect are helpful for generating better novel humorous texts in human evaluation, and that our natural language explanations can be leveraged to improve both the accuracy and robustness of humor classifiers.

\end{abstract}

\section{Introduction}


Humor serves multiple purposes and provides numerous benefits, such as relieving anxiety, avoiding painful feelings and facilitating learning~\cite{buxman2008humor}. As a specific example of humor, the creative uses of puns, wordplay and ambiguity are important ways to come up with jokes~\cite{chiaro2006language}. Pun understanding and generation are particularly challenging tasks because they require extensive commonsense and world knowledge to compose and understand, even for humans. Despite growing interest in the area, there are limited amounts of data available in the domain of humor understanding and generation.

Existing humor datasets are usually 
only annotated with binary labels indicating whether each sentence is a joke, pun, or punchline~\cite{hasan-etal-2019-ur, weller-seppi-2019-humor, castro-etal-2018-crowd, mittal-etal-2021-think}. This is insufficient to benchmark models' ability to understand and generate novel humorous text, 
since hardly anything meaningful can be learned from such a sparse supervision signal and coarse-grained annotation.
\begin{table}[t]
\small
\resizebox{0.48\textwidth}{!}{
\begin{tabular}{@{}l|l@{}}
\toprule
\textbf{Text} & \begin{tabular}[c]{@{}l@{}}When artists dream in color it's a pigment of their \\ imagination. \end{tabular}\\ \midrule
KWD & \begin{tabular}[c]{@{}l@{}} \colorbox{yellow!50}{artists}, \colorbox{yellow!50}{dream}, \colorbox{yellow!50}{color}, \colorbox{yellow!50}{pigment}, \colorbox{yellow!50}{imagination}.\end{tabular} \\ \midrule
NLEx & \begin{tabular}[c]{@{}l@{}}Pigments are non-soluble materials often used in \\painting, and pigment sounds like figment, which is \\something that is not real but someone believes it is.\end{tabular} \\ \midrule \midrule
\textbf{Text} & \begin{tabular}[c]{@{}l@{}}The man found something to catch fish, which was \\a net gain.\end{tabular}\\ \midrule
KWD &\begin{tabular}[c]{@{}l@{}} \colorbox{yellow!50}{catch fish}, \colorbox{yellow!50}{net gain}.\end{tabular}\\ \midrule
NLEx & \begin{tabular}[c]{@{}l@{}}This is a play on words. A ``net gain'' means an\\ increase in revenue but here ``net'' refers to how a net \\is used to catch fish.\end{tabular} \\ \bottomrule
\end{tabular}
}
\caption{Two examples of annotated Keywords (KWD) and Natural Language Explanations (NLEx) for puns in our dataset. The \colorbox{yellow!50}{highlighted} texts are annotated keywords that contribute to making the text funny.}
\vspace{-1em}
\label{tab:table1}
\end{table}

\begin{table*}[]
\small\centering
\resizebox{\textwidth}{!}{
\begin{tabular}{@{}l|l|l@{}}
\toprule
\textbf{Text} & Be True to your teeth, or they will be false to you. & \begin{tabular}[c]{@{}l@{}}Drinking too much of a certain potent potable \\ may require a leave of absinthe.\end{tabular} \\ \midrule
\textbf{Understandable} & {[}1, 1, 1, 1, 0{]} & {[}1, 1, 1, 1, 1{]} \\ \midrule
\textbf{Offensive/Inappropriate} & {[}0, 1, 0, 0, 0{]} & {[}0, 0, 0, 0, 0{]} \\ \midrule
\textbf{
Is a joke?} & {[}1, 0, 1, 0, 0{]} & {[}1, 1, 1, 1, 1{]} \\ \midrule
\textbf{Funniness (1-5)} & {[}2, 0, 1, 0, 0{]} & {[}3, 4, 2, 1, 2{]} \\ \midrule
\textbf{
\begin{tabular}[c]{@{}l@{}}Natural \\ Language\\ Explanation\\(NLEx)\end{tabular}} & \begin{tabular}[c]{@{}l@{}}NLEx1: Talking about being true as in being real or \\ they will be fake/false teeth.\\ NLEx2: \underline{False teeth are something people who lose} \\ 
\underline{their teeth may have}, and being true to your teeth \\ may be a way of saying take care of them otherwise \\ you'll lose them.\end{tabular} & \begin{tabular}[c]{@{}l@{}}NLEx1: It's a pun that replaces the word absence \\ with \underline{absinthe, which is notoriously strong alcohol}.\\ NLEx2: This is a play on words. Absinthe here \\ represents the liquor by the same name but is meant \\ to replace the similar-sounding ``absence''. \underline{Too much} \\ \underline{absinthe will make you ill.}\end{tabular} \\ \midrule
\textbf{
\begin{tabular}[c]{@{}l@{}} Joke keywords\\ (KWD)\end{tabular}} & \begin{tabular}[c]{@{}l@{}}KWD1: {[}``true'', ``teeth'', ``false''{]}\\ KWD2: {[}``be true'', ``teeth'', ``false to you''{]}\end{tabular} & \begin{tabular}[c]{@{}l@{}}KWD1: {[}``drinking'', ``leave of absinthe''{]}\\ KWD2: {[}``drinking too much'', ``leave of absinthe''{]}\end{tabular} \\ \bottomrule
\end{tabular}
}
\caption{Two examples with annotation fields that we collect. 
We use \underline{underline} to mark the commonsense knowledge that people need in order to understand the joke. }
\vspace{-1em}
\label{tab:annotations}
\end{table*}

To facilitate research on humor understanding and generation, we present the ExPUNations (ExPUN) dataset, in which we augment an existing dataset of puns from SemEval 2017 Task 7~\cite{miller-etal-2017-semeval} with detailed crowdsourced annotations of fine-grained funniness ratings on a Likert scale of one to five, along with keywords denoting the most distinctive words that make the text funny and natural language explanations describing why the text is funny (Table~\ref{tab:table1}). In addition, we collect annotations indicating whether a person understands the sentence, thinks it is a pun, and finds the joke offensive or inappropriate. Since these tasks are all highly subjective, we collect multiple annotations per sample, and present a detailed agreement analysis. We believe our annotations can be used in many other applications beyond pun understanding and generation, such as toxicity detection. 
%

The contributions of our work are threefold:

\begin{itemize}
  \setlength\itemsep{0.5em}
    \item We contribute extensive high-quality annotations for an existing humor dataset along multiple dimensions.\footnote{Resources will be available at: \url{https://github.com/amazon-research/expunations}} 
    \item Based on the annotations, we propose two tasks, explanation generation for pun classification and keyword-conditioned pun generation, to advance research on humor understanding and generation.
    \item We benchmark state-of-the-art NLP models on explanation generation for pun classification and keyword-conditioned pun generation. Our experiments demonstrate the benefits of utilizing natural language keywords and explanations for humor understanding and generation while highlighting several potential areas of improvement for the existing models.
\end{itemize}

\section{ExPUN Dataset}
\label{sec:dataset}
In this section, we describe our data annotation procedure, including details of the annotation fields and our assessment of the annotation quality. 

\subsection{Data Preparation}
The original SemEval 2017 Task 7 dataset~\cite{miller-etal-2017-semeval}\footnote{\url{https://alt.qcri.org/semeval2017/task7/}. The data is released under CC BY-NC 4.0 license (\url{https://creativecommons.org/licenses/by-nc/4.0/legalcode}).} contains puns that are either homographic (exploiting polysemy) or heterographic (exploiting phonological similarity to another word). The dataset also contains examples of non-pun text. We sample 1,999 text samples from SemEval 2017 Task 7 as the basis for our humor annotation.~\footnote{We sample 834 heterographic puns, 1,074 homographic puns and 91 non-puns.}

\subsection{Dataset Annotation}
The annotated fields ({$AF$}) come in the order of:
\begin{enumerate}[leftmargin=9mm]
\itemsep-.3em 
    \item [{$AF_1$}] [\emph{understandability}]: whether the annotator understands the text or not, regardless of whether they perceive it as funny. 
    \item [{$AF_2$}] [\emph{offensiveness}]: whether the annotator finds the text offensive or inappropriate.
    \item [{$AF_3$}] [\emph{joke}]: whether the annotator thinks the text is intended to be a joke.
    \item [{$AF_4$}] [\emph{funniness}]: rate the funniness on a Likert scale of 1-5, where 1 means very not funny and 5 means very funny.
    \item [{$AF_5$}] [\emph{explanation}]: explain in concise natural language about why this joke is funny. More specifically, if external or commonsense knowledge is required to understand the joke and/or its humor, the annotator should include the relevant knowledge in the explanation. If the joke is a pun or play on words, they must provide an explanation of how the play on words works.
    \item [{$AF_6$}] [\emph{joke keywords}]: pick out (as few as possible) keyword phrases from the joke that are related to the punchline/the reason the joke is funny. We emphasize that phrases should be sparse and mainly limited to content words, can be multiple words long, and the keywords should be copied verbatim from the joke.
\end{enumerate}

If an annotator rates the instance as not understandable, they will skip the rest of the annotation for that instance ({$AF_2$}-{$AF_6$}). In addition, if an annotator rates an example as not a joke, they can skip the rest of the annotation ({$AF_4$}-{$AF_6$}). Table~\ref{tab:annotations} shows two examples in our dataset. The first example has two annotators who think the text is a joke, and therefore it has two explanations. In the second instance, all annotators unanimously agree it is a joke. Here, we sample two explanations from the original five. For both instances, we use underline to highlight the external commonsense knowledge in the explanation. If the joke is a play on words, the explanation also shows how the play on words works (e.g., the second joke). We show the full annotation guidelines, including calibrating examples, in Appendix~\ref{app:annot_guidelines}.

We crowdsourced 5 annotations per sample using a professional team of 10 dedicated full-time annotators within our organization. Before starting the task, we held a kick-off meeting with the team to explain the annotation guidelines in detail. We then conducted 3 pilot rounds for calibration and iteratively met with annotators, including more details and examples to address annotator questions.~\footnote{See Appendix~\ref{app:annot_details} for more details on pilot round feedback.} Finally, we conducted 7 rounds of annotation, each with between 100-300 puns per round grouped into minibatches of 50 examples. Each sample in a minibatch was annotated by consistent subteams of 5 annotators. After receiving a completed batch of annotations, we manually examined their quality and provided feedback on any quality issues, redoing batches as necessary.

\begin{table}[]
\small
\begin{center}
\small
\resizebox{0.4\textwidth}{!}{
\begin{tabular}{ l c c c c }
\toprule
& \multicolumn{1}{c}{total} & $AF_1$ & $AF_2$ & $AF_3$ \\ \midrule 
\# samples & 1,999 & 1,795 & 65 & 1,449 \\
\midrule \midrule
\end{tabular}
}
\resizebox{0.4\textwidth}{!}{
\begin{tabular}{l r}
$AF_4$: Avg. funniness & 1.68  \\
\midrule
\multicolumn{2}{l}{$AF_5$: Explanations} \\
\quad total \# explanations & 6,650 \\
\quad avg. \# explanations/sample & 3.33 \\
\quad avg. \# tokens/expl. & 31.67 \\
\quad avg. \# sentences/expl. & 2.01 \\
\midrule
\multicolumn{2}{l}{$AF_6$: Keyword phrases} \\
\quad avg. \# tokens/keyword phrase & 1.33 \\
\quad avg. \# keyword phrases/sample & 2.09 \\
\bottomrule
\end{tabular}
}
\end{center}
\caption{\label{table:dataset-stats} Overall stats for annotation fields in ExPUN.}
\vspace{-1em}
\end{table}

\subsection{Dataset Statistics and Quality Control}

We report overall dataset statistics in Table~\ref{table:dataset-stats}. For $AF_1-AF_3$, we count the number of samples labeled positive by majority vote. For $AF_4$, we compute the average of all funniness scores, excluding blank annotations, and find that while annotators recognized most samples as jokes, they did not find them to be particularly funny. For $AF_5$ and $AF_6$, we compute lexical statistics of our explanations and keyword annotations and provide deeper analysis of these key annotation fields in Section~\ref{sec:stats}.

We report inter-annotator agreement for all annotation fields in Table~\ref{table:agreement-stats}.~\footnote{When computing agreement, we exclude the first 100 annotated samples, as these were used as a calibrating pilot.} For fields $AF_1$-$AF_4$, we compute agreement using (1) the average of Cohen's kappa scores of each annotator against the majority vote, and (2) the average Spearman correlation between each pair of annotators. We find that annotators show moderate agreement when deciding if the given text is a joke ($AF_3$), but lower agreement on the task of understanding the text ($AF_1$) as well as the much more subjective task of rating how funny a joke is ($AF_4$). We also find weak average Spearman correlation between each pair of annotations for the subjective categories of offensiveness ($AF_2$),~\footnote{See Appendix~\ref{app:offensiveness} for more details.} whether the text is a joke ($AF_3$) and joke funniness ($AF_4$). 

For the free text fields in {$AF_5$} and {$AF_6$}, we compute averaged BLEU-4~\citep{papineni-etal-2002-bleu} and METEOR~\citep{banerjee-lavie-2005-meteor} scores in a pairwise fashion. We treat each annotator's explanation (for ${AF_5}$) or list of keyword phrases joined into a string (for ${AF_6}$) as candidate text, with the remaining annotators' annotations as a set of references. We find high similarity between joke keyword annotations, suggesting that annotators identify similar spans of keyword phrases, and a lower degree of similarity between pun explanations.

\subsection{Dataset Analysis}\label{sec:stats}


\begin{table}[]
\small
\begin{center}
\small
\begin{tabular}{ l | c | c | c | c}
\toprule
 Annotation Field & $\kappa$ & $\rho$ & BLEU & MET. \\ \midrule
$AF_1$: Understand (0/1) & 0.40 & 0.16 & -& -\\
$AF_2$: Offensive (0/1) & 0.16 & 0.34 & -& -\\
$AF_3$: Joke (0/1) & 0.58 & 0.32 & -& -\\
$AF_4$: Funny (1-5) & 0.41 & 0.30 & -& -\\
$AF_5$: Explain (Text) & - & - & 0.18 & 0.30 \\
$AF_6$: Keywords (Text) & - & - & 0.58 & 0.74 \\
\bottomrule
\end{tabular}
\end{center}
\vspace{-.5em}
\caption{\label{table:agreement-stats}Agreement stats for annotated fields in the ExPUN dataset. We report averaged Cohen's $\kappa$ and Spearman's $\rho$ for numeric ratings ($AF_1-AF_4$), and averaged BLEU-4 and METEOR for text fields ($AF_5-AF_6$).}
\vspace{-1em}
\end{table}

\paragraph{Explanations.} 

\begin{figure}
    \centering
    \small
     \begin{subfigure}[t] {0.235\textwidth}
            \includegraphics[trim=0cm 0cm 0.1cm 0cm, clip, width=\linewidth]{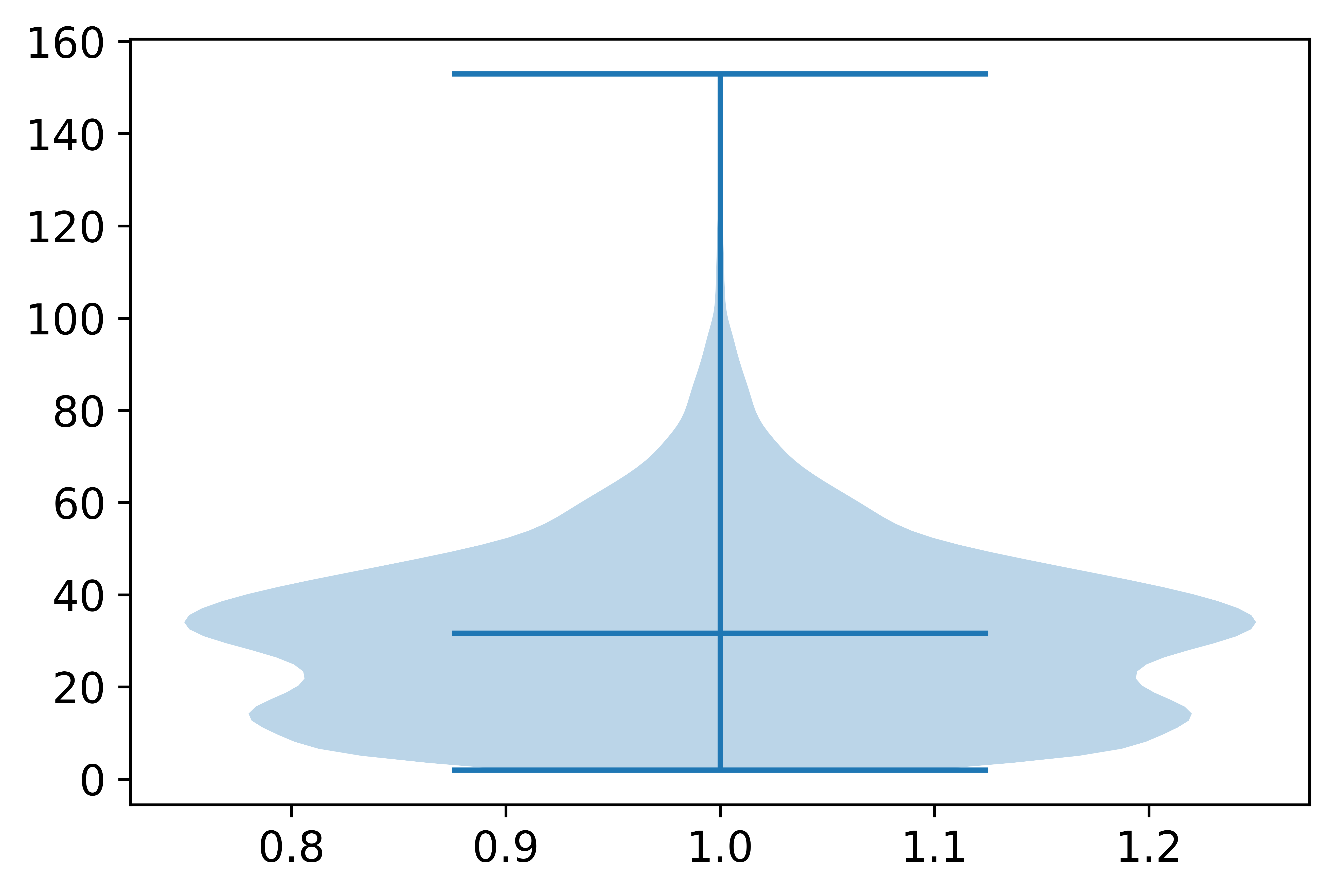}
            \caption{\small Tokens/explanation}
            \label{fig:tokens-per-expl}
    \end{subfigure}
    \begin{subfigure}[t] {0.235\textwidth}
            \includegraphics[trim=0.1cm 0cm 0cm 0cm, clip,width=\linewidth]{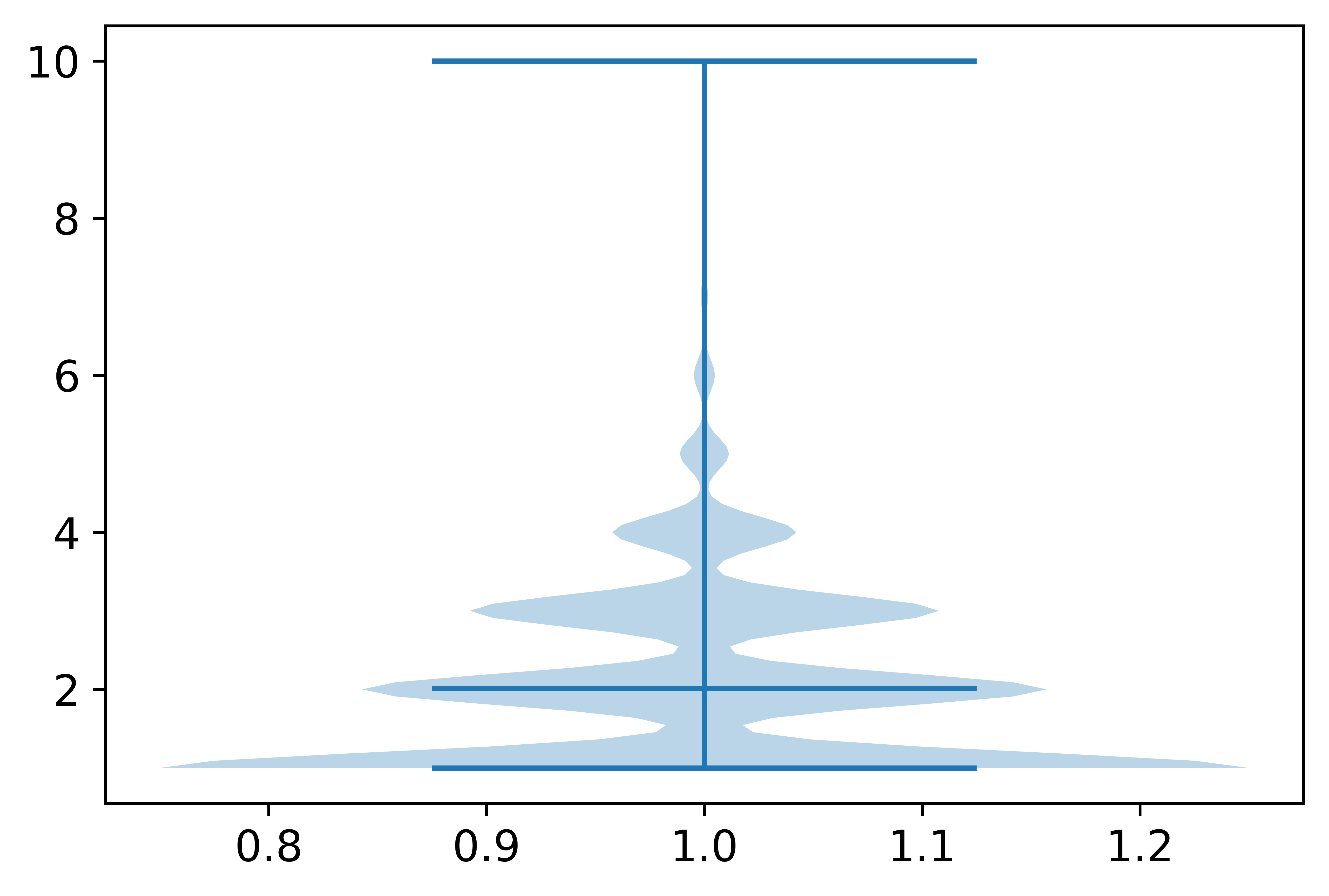}
            \caption{\small Sentences/explanation}
            \label{fig:num-sens-per-expl}
    \end{subfigure}
    \begin{subfigure}[t] {0.235\textwidth}
            \includegraphics[trim=0.1cm 0cm 0.1cm 0cm, clip, width=\linewidth]{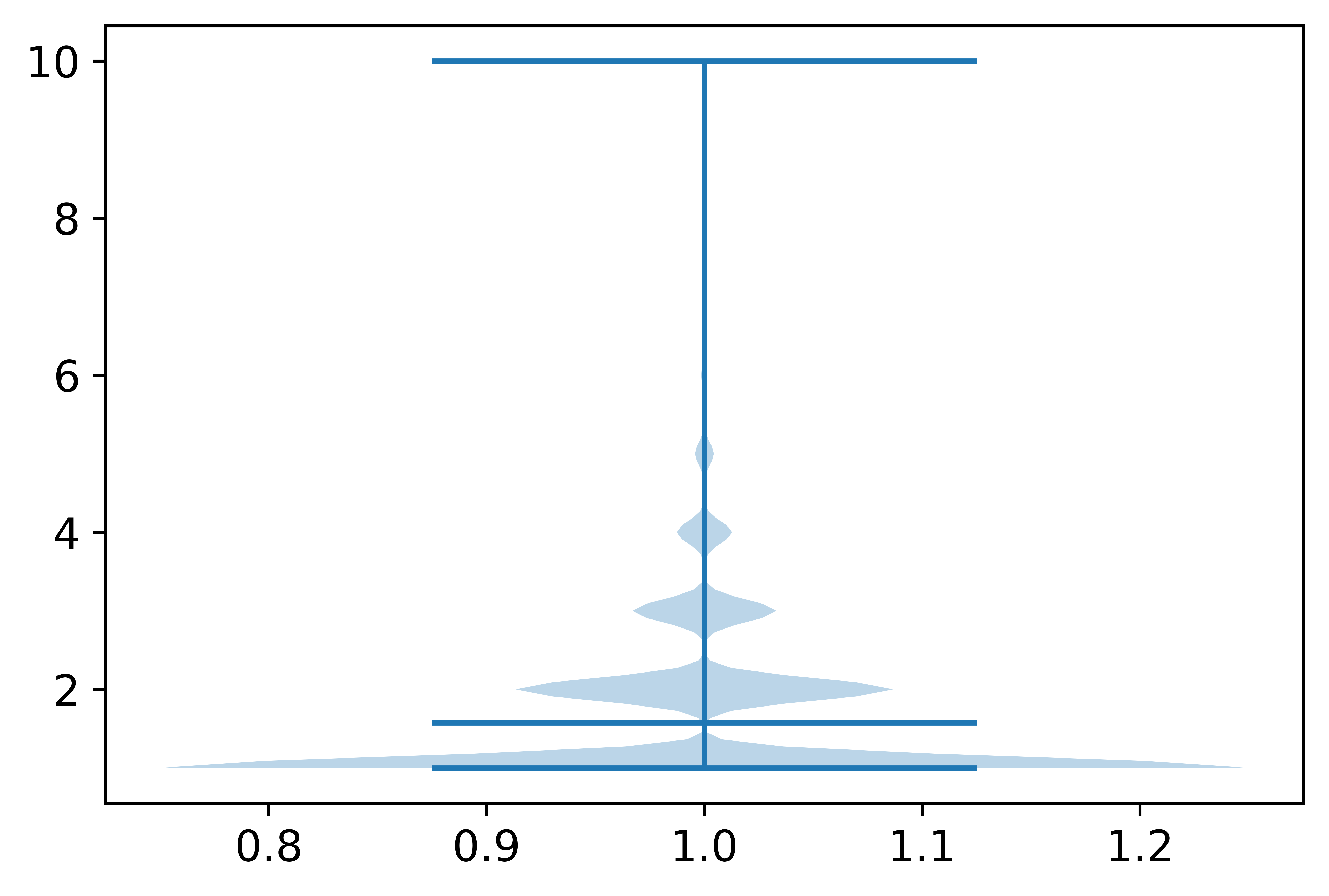}
            \caption{\small Tokens/keyword phrase}
            \label{fig:tokens-per-kwp}
    \end{subfigure} 
    \begin{subfigure}[t] {0.235\textwidth}
            \includegraphics[trim=0.1cm 0cm 0cm 0cm, clip,width=\linewidth]{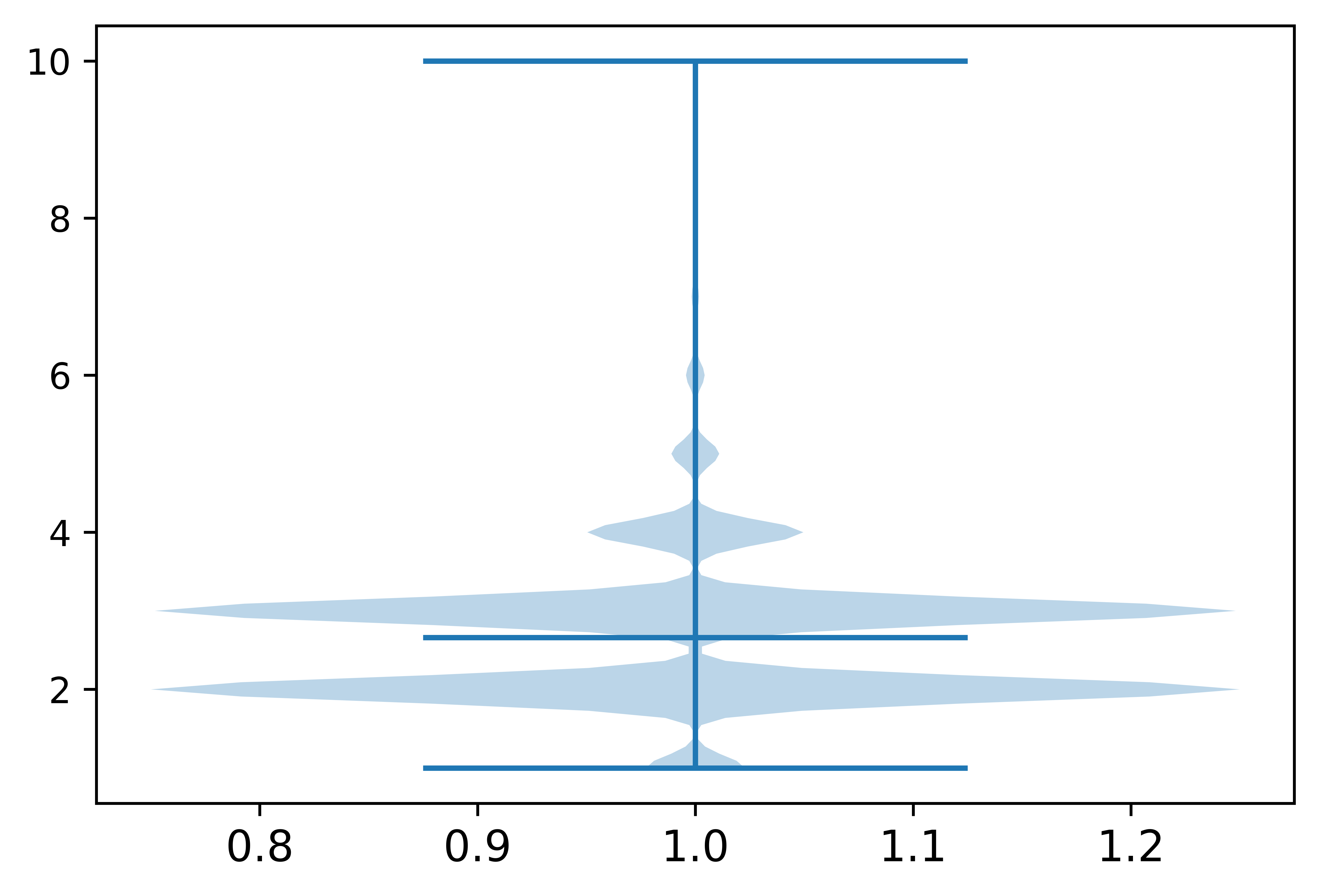}
            \caption{\small Keyword phrases/sample}
            \label{fig:kwps-per-sample}
    \end{subfigure}
    \caption{Distributions of (a) number of tokens and (b) number of sentences in explanations ($AF_5$), (c) tokens in keyword phrases ($AF_6$), and (d) keyword phrases per sample. Horizontal lines are used to show the min, mean, and max values for each distribution.}
    \vspace{-1em}
    \label{fig:dataset_violins}
\end{figure}

As seen in Figures~\ref{fig:tokens-per-expl} and~\ref{fig:num-sens-per-expl}, on average, samples are annotated with multiple explanations, and the explanations are lengthy, spanning multiple sentences, and lexically diverse (14,748 token vocabulary size, with 210,580 tokens overall). 
Figure \ref{fig:most_frequent} in Appendix \ref{app:explanations} shows the distribution of the top 50 most frequent content-words in our explanations. The frequent use of {\it usually} and {\it often} indicate the explanation of commonsense knowledge, e.g., {\it thunder and lightning are usually present in a weather storm} or {\it ``pain'' means physical discomfort often felt by a hospital patient.} The most frequent words, {\it means} and {\it word}, indicate that annotators frequently provide word sense information as part of their explanations, while {\it sounds} frequently appears in explanations of heterographic puns. Each of these most frequent words comprise less than 2.8\% of all tokens in the explanations, illustrating the rich diversity of our corpus.~\footnote{We show an analysis of highly-frequent explanation templates, as well as unique and highly-informative templates, in Appendix \ref{app:explanations}.}

\paragraph{Keywords.} 
As seen in Figures~\ref{fig:tokens-per-kwp} and~\ref{fig:kwps-per-sample}, on average, keyword phrases in ExPUN, which are derived from the original puns, are short and sparse (5,497 token vocabulary size, with 27,820 tokens overall). 
This follows from our guidelines to annotate keywords concisely, focusing mainly on content words that are essential to understanding the joke. Table~\ref{tab:keywords} shows two examples of pun keyword annotations in our dataset that showcase different annotation styles among annotators. For instance, one annotator may tend to select wordy keyword phrases that introduce unnecessary tokens, while another may omit salient keywords that other annotators mention. Aggregating these annotations among annotators to construct a single ground truth set of keyword phrases is therefore challenging because of differing annotation styles. The problem of merging keywords is further complicated because the keywords from different annotators are often not aligned well, as different annotators may annotate varying numbers of keyword phrases and different spans. Taking these considerations into account, we propose a keyword aggregation algorithm to address these issues and construct a single set of aggregated keywords per sample.

\paragraph{Keywords Aggregation.} 
Algorithm~\ref{alg:template} in Appendix \ref{app:aggregated-keyword} describes our keyword aggregation method. The algorithm aims to generate a comprehensive list of concise keywords for each sample. First, we compute a reliability score for each annotation, defined as the average of (\# keyword phrases$-$\# average tokens in each keyword phrase). The higher the score, the more comprehensive and concise the keywords from an annotator should be. We choose the annotator with the highest score to be the \textit{anchor}. 
We note, however, that keyword annotations are not always error-free; e.g., in the first example of Table~\ref{tab:keywords}, $w_{4}$ has an incorrect word (\textit{fancy chairs} instead of \textit{royal chairs}). Therefore, for each keyword phrase, we compute the fuzzy matching score
between the anchor's annotation with the rest of annotators' annotations. For each annotator, we keep the keyword phrase that has the highest fuzzy matching score with the anchor annotator's, with a minimum threshold score of 60.~\footnote{This is empirically determined.} This process produces a filtered keyword list where each of the remaining keyword phrases look similar to the anchor's. Then, we compute the average fuzzy matching score between the anchor's keyword phrase and each element in the filtered keyword list. We then choose the annotator with the second-highest reliability score to be the anchor, and repeat the above process. Finally, by choosing the resulting keyword phrases that attain the maximum average fuzzy matching score between the first and second anchors, we get the final aggregated keywords for this instance.

\begin{table}[]
\small
\resizebox{0.48\textwidth}{!}{
\begin{tabular}{@{}l|l|l@{}} \toprule
\textbf{} & \textbf{\begin{tabular}[c]{@{}l@{}}Royal chairs are \\ rarely throne out.\end{tabular}} & \textbf{\begin{tabular}[c]{@{}l@{}}She didn't marry the \\ gardener. Too rough \\ around the hedges.\end{tabular}} \\ \midrule
\textbf{$w_1$} & \begin{tabular}[c]{@{}l@{}}{[}Royal chairs, \\ throne out{]}\end{tabular} & \begin{tabular}[c]{@{}l@{}}{[}didn't marry the gardener, \\ too rough around the hedges{]}\end{tabular} \\ \midrule
\textbf{$w_2$} & \begin{tabular}[c]{@{}l@{}}{[}Royal chairs, \\ throne out{]}\end{tabular} & \begin{tabular}[c]{@{}l@{}}{[}didn't marry the gardener, \\ rough around the hedges{]}\end{tabular} \\ \midrule
\textbf{$w_3$} & \begin{tabular}[c]{@{}l@{}}{[}Royal chairs, \\ rarely throne out{]}\end{tabular} & \begin{tabular}[c]{@{}l@{}}{[}didn't marry the gardener, \\ rough around the hedges{]}\end{tabular} \\ \midrule
\textbf{$w_4$} & \begin{tabular}[c]{@{}l@{}}{[}fancy chairs, \\ throne{]}\end{tabular} & {[}gardener, rough, hedges{]} \\ \midrule
\textbf{$w_5$} & \begin{tabular}[c]{@{}l@{}}{[}Royal chairs, \\ throne{]}\end{tabular} & \begin{tabular}[c]{@{}l@{}}{[}gardener, \\ rough around the hedges{]}\end{tabular} \\ \midrule\midrule
 \textbf{$w_A$} & \begin{tabular}[c]{@{}l@{}}{[}royal chairs, \\ throne out{]}\end{tabular} & {[}gardener, rough, hedges{]} \\ \bottomrule
\end{tabular}
}
\caption{Keyword annotations from different workers. $w_A$ shows aggregated keywords from our algorithm.}
\vspace{-1em}
\label{tab:keywords}
\end{table}
\begin{figure*}
    \centering
     \begin{subfigure}[t] {0.33\textwidth}
            \includegraphics[trim=0.1cm 0.0cm 0.0cm 0.0cm, clip, width=\linewidth]{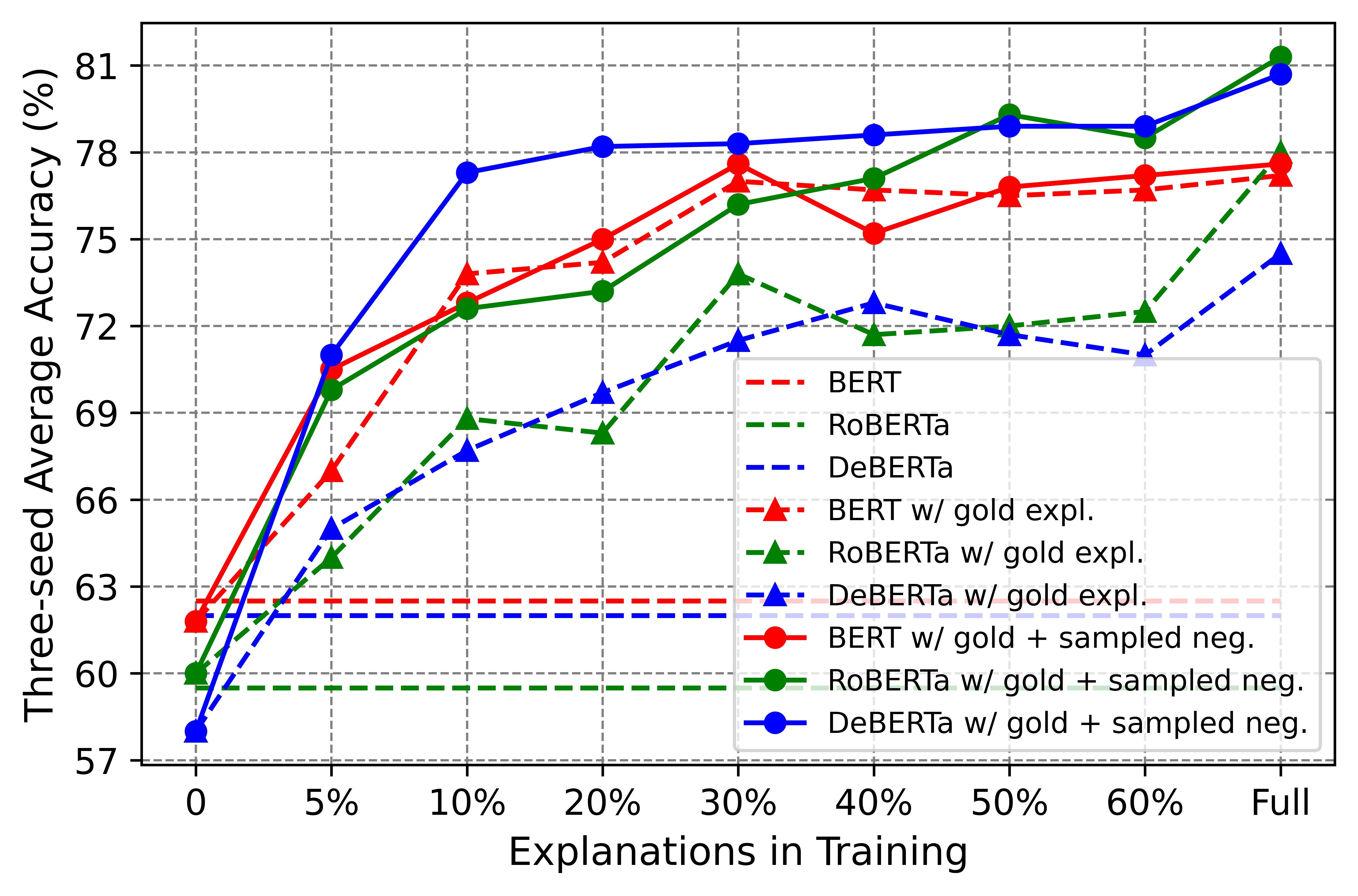}
            \caption{Gold explanations during test.}
            \label{fig:gold-inference}
    \end{subfigure}
    \begin{subfigure}[t] {0.32\textwidth}
            \includegraphics[trim=0.1cm 0cm 0.1cm 0cm, clip, width=\linewidth]{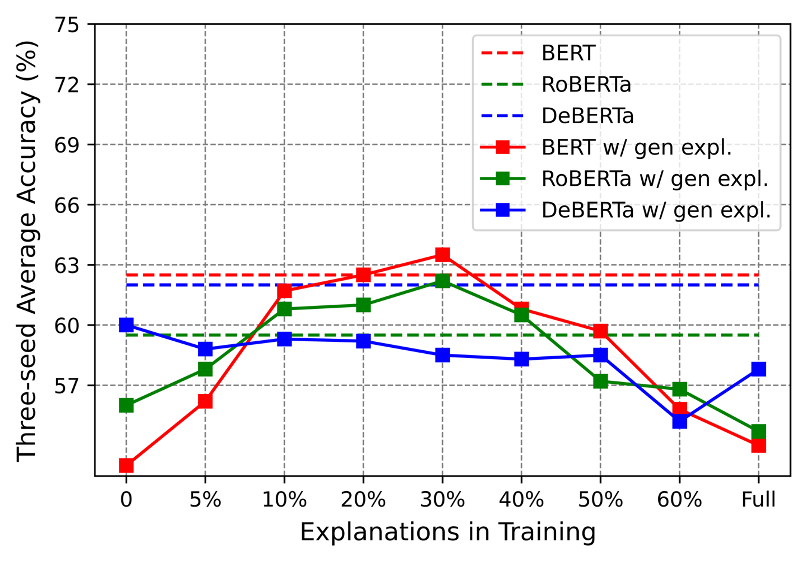}
            \caption{Generated explanations during test.}
            \label{fig:generated-e}
    \end{subfigure} 
    \begin{subfigure}[t] {0.32\textwidth}
            \includegraphics[trim=0.1cm 0cm 0cm 0cm, clip,width=\linewidth]{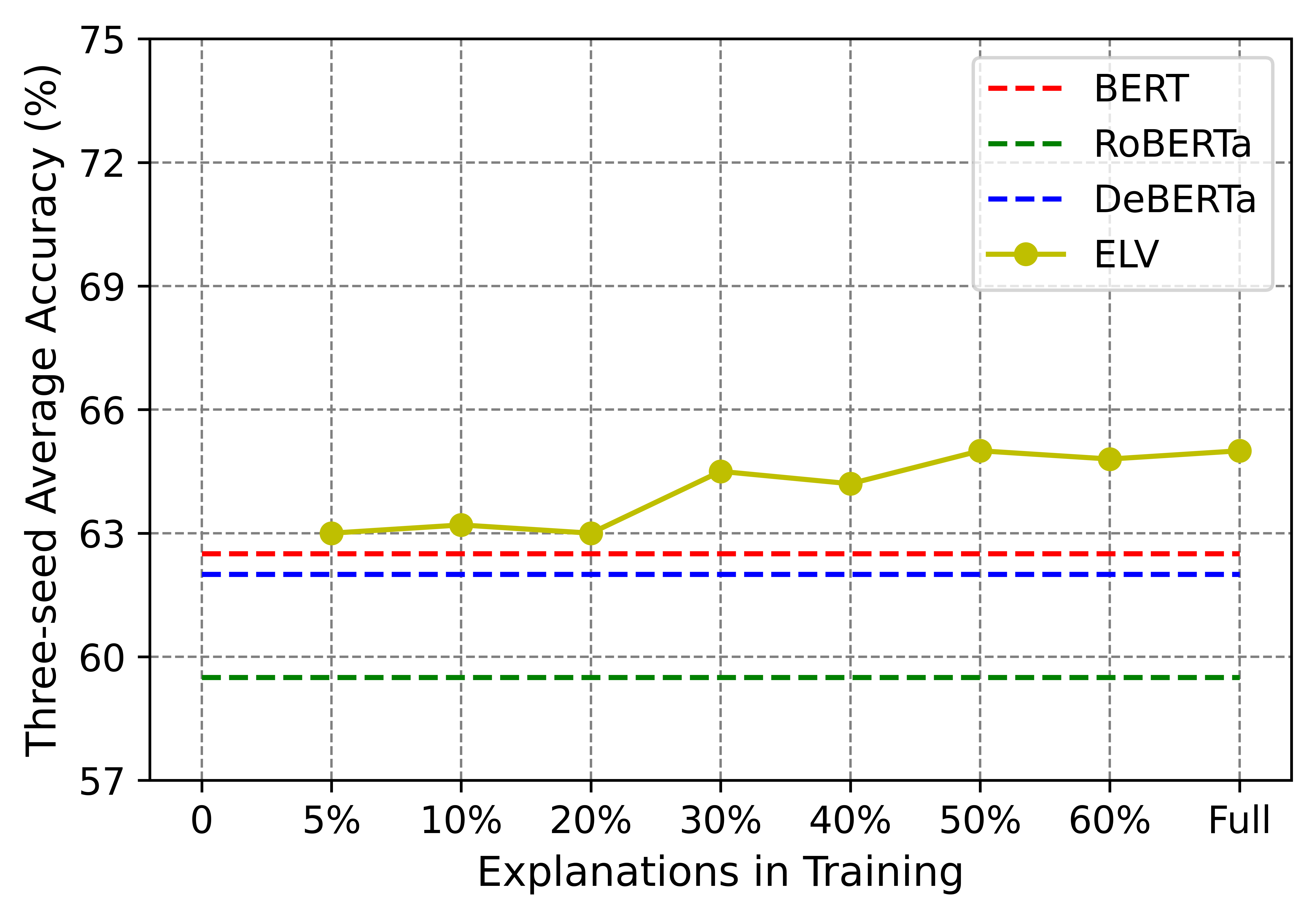}
            \caption{ELV model.}
            \label{fig:ELV}
    \end{subfigure}
    \caption{The impact of using human-written (\ref{fig:gold-inference}) and model-generated explanations (\ref{fig:generated-e} and~\ref{fig:ELV}) vs. no explanations (constant dotted lines) on pun classification accuracy. All reported numbers are computed with three-seed average. For each data point, we train a model on the full dataset, but only provide explanations for a given percentage, as shown on the x-axis.}
    \vspace{-1em}
    \label{fig:models}
\end{figure*}

\section{Experiments}

With the collected annotations, we propose two new tasks, pun explanation and keyword conditioned pun generation, to showcase novel tasks that our dataset uniquely enables and push the frontiers of NLU and NLG for humor. 
Note that the rich annotations in ExPUN can also enable many other interesting tasks such us pun keywords extraction, fine-grained funniness prediction, and others. However, we prioritize NLG tasks as they are relatively under-explored compared to NLU tasks. 
In this section, we benchmark current state-of-the-art models' performance on the proposed tasks.
\subsection{Pun Explanation}
The task of pun explanation takes a pun sentence as input and outputs a natural language explanation of why the pun is funny. This requires extensive understanding of background and commonsense knowledge. We hypothesize that existing NLP models would struggle to generate high-quality explanations for puns. On the other hand, high-quality explanations can improve humor understanding, and thus help tasks such as humor classification.

Formally, given text $T$, our target is to generate an explanation $E_T$ of why $T$ is funny. Additionally, we use the explanations to support the task of pun classification, where, given $T$ (and optionally an explanation $E_T$), we output whether $T$ is a joke. 

\paragraph{Data Preparation.} For each data sample, we use the longest human-written explanation from ExPUN ($AF_5$), substituting in the pun text if no explanations exist.~\footnote{Only 168 samples have no annotated explanations.} For pun classification, we assign output labels using the majority vote of $AF_3$ (is a joke). For both tasks, we split our dataset into 1,699/100/200 for train/dev/test. Dev and test contain an equal distribution jokes to non-jokes, while training contains 1,299 jokes and 400 non-jokes.

\paragraph{Evaluation Metrics.} We do not report lexical overlap metrics as our primary evaluation metric for generated explanations because these are not suited for measuring plausibility~\cite{camburu2018snli, Kayser2021eViLAD, clinciu-etal-2021-study} or faithfulness of explanations~\cite{jacovi-goldberg-2020-towards}. Rather, we follow prior work and use the ``\textit{simulatability score}'' metric from \citet{wiegreffe-etal-2021-measuring} to measure explanation quality from the lens of usability of the explanation. It reflects the utility of explanations by measuring the improvement in task performance when explanations are provided as additional input vs. when they are not: $\textrm{acc(IE}\rightarrow\textrm{O}) - \textrm{acc(I}\rightarrow\textrm{O})$, where $\textrm{I}$ denotes the input text, $\textrm{E}$ is the explanation and $\textrm{O}$ is the classification of whether $\textrm{I}$ is a joke. We evaluate {how useful explanations can be} by measuring the performance increase of $\textrm{acc(IE} \rightarrow \textrm{O})$ as we increase the ratio of samples with explanations in the training data, and report $\textrm{acc(I} \rightarrow \textrm{O})$ as a constant baseline that uses no explanations.

\paragraph{Models.} We use the following model variations:~\footnote{Further experimental details in Appendix \ref{sec:appendix-classifier}.} 
    
    \medskip 
    \noindent \ul{\textit{No explanations.}} As a baseline, we finetune BERT-base~\cite{devlin-etal-2019-bert}, RoBERTa-base~\cite{liu2019roberta} and DeBERTa-base~\cite{he2020deberta} to classify whether the given text is a joke without any explanations in the input. 
    
    \medskip 
    \noindent \ul{\textit{Gold explanations.}} To find the upper bound of {how useful explanations can be}, we augment the input to the above baseline models with gold human-annotated explanations in both training and testing. {The majority of non-punny examples (identified as unfunny by majority vote and thus labeled as unfunny) contain at least one explanation from an annotator who marked it as funny. In these cases, we use any provided explanations as $E$, both in training and in testing with gold explanations. Otherwise, to construct training examples that have no annotated explanations, or where explanations are held out, we try two variants: (1) representing the missing explanation as an empty string (\textit{``w/ gold expl.''}), or (2) randomly sampling a negative explanation from another annotated example to use as input (\textit{``w/ gold + sampled neg.''}).}
    
    \medskip 
    \noindent \ul{\textit{Generated explanations.}} Following previous work on explanation generation~\cite{wiegreffe-etal-2021-measuring}, we first finetune a T5~\cite{t5} model to generate pun explanations given pun sentences as input. For text that contains no annotated explanations, we use the pun sentence itself as the output explanation. We then use gold human-annotated explanations to train and T5-generated explanations to test the explanation-augmented classification models. 
    
    \medskip 
    \noindent \ul{\textit{ELV}}~\cite{Zhou2020TowardsIN}. ELV is a probabilistic framework for text classification where natural language {\bf E}xplanations are treated as {\bf L}atent {\bf V}ariables. Two modules, an explanation generation module and an explanation-augmented prediction module are jointly trained using a variational EM framework. As another baseline, we train an ELV model for pun classification using the ExPUN dataset.

\paragraph{Results.}
We show our results on the pun classification task in  Figure~\ref{fig:models}. Baseline performance of the \textit{no explanations} models are shown using constant dotted lines. Figure~\ref{fig:gold-inference} shows the upper bound of performance improvement when models are provided with \textit{gold explanations}, indicating that human-written explanations are useful for this task, and that including more gold explanations in training data generally helps. {In particular, adding randomly-sampled negative explanations (``\textit{w/ gold + sampled neg.}'') further improves the classification accuracy, showing the utility of our collected explanations in improving model performance.} However, Figure~\ref{fig:generated-e} shows that using \textit{generated explanations} at test time does not help to improve classification accuracy. Using the more carefully-designed ELV framework to jointly train the generation and classification modules shows improvement in classification accuracy (Figure~\ref{fig:ELV}); however, qualitative analysis of the ELV explanations showed that many generated outputs are not fluent natural language, suggesting that performance improvements may stem more from modeling improvements as opposed to explanations. Given the huge improvements we see when incorporating gold explanations during test, we note explanations are clearly highly valuable if the quality of generated explanations can be improved.

Table \ref{table:generated-explanations} shows examples of T5-generated explanations for given puns. Qualitative analysis shows that generated explanations often identify the relevant pun word, and can include somewhat accurate word sense information for one sense of the pun. However, the model usually fails to explain the alternate word sense and its relation, which is crucial to understanding the wordplay. The model especially fails to explain phonological similarity in heterographic puns; e.g., in the first three examples, explanations fail to mention alternate words \textit{carry}, \textit{whet} and \textit{humor}. For both pun types, our model can devolve into repetitively copying words from the input.
Our results exhibit the challenge of generating good pun explanations and that high-quality explanations are useful for understanding humor.

\begin{table}[t]
\small
\begin{center}
\resizebox{0.48\textwidth}{!}{
\small
\begin{tabular}{ l l }
\toprule
 \textbf{Input Pun} & \textbf{Generated Explanation} \\ \midrule
 \begin{tabular}[c]{@{}l@{}}My name is \\ \underline{Cary.} I'm a \\ porter.\end{tabular} & \begin{tabular}[c]{@{}l@{}}The joke is a pun on the word \\ ``cary''. A porter is someone who \\ transports goods.\end{tabular} \\ \midrule
 \begin{tabular}[c]{@{}l@{}}Fishers often \\ \underline{wet} their \\ appetite.\end{tabular} & \begin{tabular}[c]{@{}l@{}}This is a play on words. The word \\ ``wet'' means to wet your appetite, \\ which is a characteristic of fish.\end{tabular} \\ \midrule
  \begin{tabular}[c]{@{}l@{}}A gossip is \\ someone with \\ a great sense \\ of \underline{rumor}.\end{tabular} & \begin{tabular}[c]{@{}l@{}}This is a play on words. The word \\ ``rumor'' sounds like ``rumor''. A \\ gossip is someone who has a \\ great sense of rumor. \end{tabular} \\ \midrule
 \begin{tabular}[c]{@{}l@{}}Oil executives \\ are always \\ using \underline{crude} \\ language.\end{tabular} & \begin{tabular}[c]{@{}l@{}}The joke is a pun on the word \\ ``crude''. Crude language is used \\ to describe crude oil, which is a \\ type of petroleum product.\end{tabular} \\ \midrule
  \begin{tabular}[c]{@{}l@{}}Please mix me \\ a martini, said \\ Tom, \underline{dryly}. \end{tabular} & \begin{tabular}[c]{@{}l@{}}This is a play on words. The word \\ ``dryly'' means dryly, but ``dryly'' \\ sounds like ``dryly''.  \end{tabular} \\ \bottomrule
\end{tabular}}
\end{center}
\caption{\label{table:generated-explanations} Pun explanations generated by the T5 model. We use \underline{underline} to indicate the pun word in the input.}
\vspace{-1em}
\end{table}

\subsection{Keyword-Conditioned Pun Generation}
The task of keyword-conditioned pun generation takes human-annotated pun keywords as input and produces novel puns as output. This benchmarks models' capability to draw connections among words to generate novel fluent, sensible, and humorous texts. 
This is a challenging task with many downstream applications, such as context-situated humor generation, a task that involves generating humorous text in a given situation or context. In this case, input keywords can come from conversational context (e.g., chatbot dialogues) or narrative context (e.g., creative short stories).

More formally, we take as input keywords $K$, the pun word $p_w$ and alternate pun word $a_w$,~\footnote{$p_w=a_w$ for homographic puns.} and produce novel and fluent puns that incorporate the keywords.~\footnote{{We refer to ``fluent puns'' primarily in the context of the pun generation task, since generating fluent natural language realizations is often non-trivial, particularly in the case of controllable language generation tasks such as ours.}} Optionally, we also include pun word sense annotations $S_{p_w}$ and $S_{a_w}$ from the original SemEval 2017 Task 7 annotations.

\paragraph{Data Preparation.} For this task, we limit our data to samples that contain both (1) annotated human keywords $K$ from ExPUN ($AF_6$), and (2) pun word sense annotations $S_{p_w}$ and $S_{a_w}$ from SemEval 2017 Task 7. There are 1,482 such samples that have both annotations, from which we reserve 100 as test data and use the rest for model training. To construct input human-annotated keywords for this task, we aggregate keywords for each sample using the method described in Section~\ref{sec:stats}. Additionally, we evaluate the effect of finetuning on automatically-extracted keywords instead of human-annotated keywords by automatically extracting keywords for each sample by running the RAKE~\cite{rakekeywords} algorithm on the pun text.

\paragraph{Evaluation Metrics.} We use both automatic metrics and human evaluation to evaluate the quality of generated puns. For automatic evaluation, we calculate word incorporation rate for both pun words and keywords, which measure the model's ability to incorporate all input keywords. Additionally, we run human evaluation using Amazon Mechanical Turk, in which we asked Turkers to label whether or not a given generated pun was successful.~\footnote{Turkers had to pass a qualifier by correctly labeling $>=80\%$ of 20 samples that we manually annotated. Success is defined as whether the text supports both senses of the pun word. We measure inter-annotator agreement among 3 annotators using Fleiss' kappa ($\kappa=0.49$), showing moderate agreement.}

\paragraph{Models.} We use the following models:
    
    \medskip
    \noindent \ul{\textit{AmbiPun}}~\citep{mittal2022ambipun}. We use the current state-of-the-art homographic pun generation model, AmbiPun, with no further finetuning. We follow the AmbiPun prompt format: ``generate sentence: $K$, $p_w$, $a_w$''.
    
    \medskip

    \noindent \ul{\textit{Finetuned T5}} (T5$_\textrm{FT}$). We finetune T5-base on ExPUN using input prompt ``generate a pun that situated in $K$, using the word $p_w$, $p_w$ means $S_{p_w}$, $a_w$ means $S_{a_w}$.'' The output is the pun itself.~\footnote{Further experimental details in Appendix \ref{sec:appendix-t5}.}
    
    \medskip
    \noindent \ul{\textit{Finetuned T5 with pretraining}} (T5$_\textrm{PT+FT}$). To increase the model's ability to incorporate keywords, we pretrain T5 on non-pun text.  For a given pun word, we first extract 200 sentences that contain the pun word from BookCorpus~\citep{bookcorpus}, then use RAKE to automatically extract keywords for each sentence. {We construct examples where inputs are automatically extracted keywords, and outputs are sentences from BookCorpus including pun words.} We pretrain a T5 model with this data before finetuning it on ExPUN.


\begin{table}[]
\small
\small
\begin{tabular}{@{}llcccc@{}} \toprule\multicolumn{1}{c}{\textbf{Key-}}
 &  & \multicolumn{3}{c}{\textbf{Word Incorp. \%}} & \multicolumn{1}{l}{\textbf{Success}} \\ 
\multicolumn{1}{l}{\textbf{words}} & \textbf{Model} & \multicolumn{1}{c}{\textbf{$p_w$}} & \multicolumn{1}{c}{\textbf{$K$}} & \multicolumn{1}{c}{both} & \multicolumn{1}{l}{\textbf{Rate \%}} \\ \midrule
\multicolumn{1}{l}{\multirow{2}{*}{\begin{tabular}[c]{@{}l@{}}RAKE\end{tabular}}} & T5$_\textrm{FT}$ & 90.0 & 76.4 & 80.2 & 35.0 \\ 
\multicolumn{1}{l}{} & T5$_\textrm{PT+FT}$ & \textbf{99.0} & 72.9 & 81.2 & 54.0 \\ 
\midrule
\multicolumn{1}{l}{\multirow{3}{*}{\begin{tabular}[c]{@{}l@{}}ExPUN\end{tabular}}} & AmbiPun 
& \textbf{99.0} & \textbf{92.1} & \textbf{94.4} & 51.0 \\ 
\multicolumn{1}{l}{} & T5$_\textrm{FT}$ & 58.0 & 80.3 & 72.3 & 40.0 \\ 
\multicolumn{1}{l}{} & T5$_\textrm{PT+FT}$ & 93.0 & 80.2 & 83.5 & \textbf{77.0} \\ 

 \bottomrule
\end{tabular}
\caption{Automatic (Word Incorporation Rate) and human evaluation (Success \%) of puns generated by models finetuned using automatically-extracted (RAKE) and human-annotated (ExPUN) keywords (with AmbiPun baseline \cite{mittal2022ambipun}). \texttt{PT} stands for Pre-Training and \texttt{FT} stands for Fine-Tuning. {Both T5$_\textrm{PT+FT}$ models finetuned with RAKE-based keywords or ExPUN-based keywords use RAKE-based keywords during pretraining.}
}
\vspace{-1em}
\label{tab:pun-gen}
\end{table}
\paragraph{Results.}

Table~\ref{tab:pun-gen} shows results of our pun generation models. While the AmbiPun baseline achieves superior word incorporation performance, our T5$_\textrm{PT+FT}$ model finetuned using ExPUN keywords generates successful puns at a higher rate, showing the value of training on our dataset. Furthermore, while pun word incorporation is improved by pretraining on outside sources using RAKE keywords, using automatically-extracted keywords when training on in-domain pun text does not translate to more successful puns. Instead, models finetuned with the more carefully-selected, human-annotated ExPUN keywords generate puns relatively more successfully than their RAKE-trained counterparts.

Table~\ref{table:generated-puns} shows examples of generated puns from our ExPUN-T5$_\textrm{PT+FT}$ model. The model is able to generate both homographic and heterographic puns somewhat coherently using one of the pun word senses. However, while some puns are successful, Rows 3 and 6 show some ways our model can struggle to generate the respective pun types: it does not always incorporate the alternate word sense in a clever or meaningful way, and can stitch copied input keywords together into incoherent sentences. Our results show pun generation is a very challenging task, and that careful selection of pun keywords and a deeper understanding of humor in wordplay is essential for generating puns successfully. 


\begin{table}[t]
\small

\begin{center}
\resizebox{0.48\textwidth}{!}{
\small
\begin{tabular}{ l l l l }
\toprule
\# & \multicolumn{1}{c}{$p_w, a_w$} & \multicolumn{1}{c}{$K$} & \textbf{Generated Pun} \\ \midrule 
1 & \begin{tabular}[c]{@{}l@{}}solution/ \\ solution\end{tabular} & \begin{tabular}[c]{@{}l@{}}scientist, \\ problem, \\ liquid \\ chemicals\end{tabular} & \begin{tabular}[c]{@{}l@{}}A liquid chemicals \\ scientist has a problem \\ with a solution.\end{tabular} \\ \midrule
2& \begin{tabular}[c]{@{}l@{}}makeup/ \\ makeup\end{tabular} & \begin{tabular}[c]{@{}l@{}}class, \\ beauty \\ school\end{tabular} & \begin{tabular}[c]{@{}l@{}}A beauty school class \\ was cancelled because \\ of a lack of makeup.\end{tabular} \\ \midrule
3&  \begin{tabular}[c]{@{}l@{}}charges/ \\ charges\end{tabular} & \begin{tabular}[c]{@{}l@{}}farmer, \\ bull\end{tabular} & \begin{tabular}[c]{@{}l@{}}The farmer, the bull, \\ had to pay the charges.\end{tabular} \\ \midrule
4& \begin{tabular}[c]{@{}l@{}}fission/ \\ fishing\end{tabular} & \begin{tabular}[c]{@{}l@{}}nuclear \\ physicist, \\ vacation,  trip\end{tabular} & \begin{tabular}[c]{@{}l@{}}The nuclear physicist \\ took a trip to the Bahamas \\ for his fission vacation.\end{tabular} \\ \midrule
 5 & \begin{tabular}[c]{@{}l@{}}fare/ \\ fair\end{tabular} & \begin{tabular}[c]{@{}l@{}}carnival, \\ county \end{tabular} & \begin{tabular}[c]{@{}l@{}}The carnival in the \\ county was a fare event.\end{tabular} \\ \midrule
 6& \begin{tabular}[c]{@{}l@{}}vault/ \\ fault\end{tabular} & \begin{tabular}[c]{@{}l@{}}bankers, \\ generous \end{tabular} & \begin{tabular}[c]{@{}l@{}}OLD BACHERS never \\ die they just become very \\ generous. They have a \\ vault fault.\end{tabular} \\
\bottomrule
\end{tabular}}
\end{center}
\caption{\label{table:generated-puns}Examples of input pun words and keywords and the resulting generated puns. We show examples of both homographic and heterographic generated puns.}
\vspace{-1em}
\end{table}
\section{Related Work}
In this work, we contribute annotations for a humor dataset as well as two humor-related generation tasks. The work is broadly related to pun generation, pun detection, explanation generation, and humor generation. We briefly summarize works in these directions.
\paragraph{Pun generation.}
Many of the previous works on pun generation have focused on phonological or syntactic patterns rather than semantic patterns \cite{miller-gurevych-2015-automatic, hong-ong-2009-automatically, petrovic-matthews-2013-unsupervised,inproceedings2000}, thus lacking flexibility.  \citet{he2019pun} make use of local-global surprisal principle to generate homophonic puns and \citet{yu-etal-2020-homophonic} uses constrained lexical rewriting for the same task. \citet{hashimoto2018retrieve} use a retrieve and edit approach to generate homographic puns and \citet{yu2018neural,luo2019pun} propose complex neural model architectures such as constrained language model and GAN. 
~\citet{mittal2022ambipun} generate homographic puns given a polyseme and try to incorporate the multiple senses of the polyseme. 
~\citet{tian2022unified} proposed a unified framework to generate both homographic and homophonic puns leveraging humor principles.  
Our keyword-conditioned pun generation task encourages models to focus more on the linguistic structures via pun keywords as we observe that human-extracted keywords usually reflect the ambiguity and distinctiveness principles as discussed in \citet{kao2016computational}. 
The keyword-conditioned pun generation setup can also facilitate more engaging pun generation scenarios such as context-situated pun generation~\cite{sun2022context}.

\paragraph{Humor generation.} 
With the recent advent of diverse datasets \cite{hasan-etal-2019-ur,mittal-etal-2021-think,yang-etal-2021-choral}, it has become easier to detect and generate humor. While large pre-trained models have become fairly successful at detection, humor generation still remains an unsolved problem. Therefore, humor generation is usually studied in a specific settings.~\citet{petrovic-matthews-2013-unsupervised} generates jokes of the type 'I like my X like I like my Y, Z'. \citet{garimella-etal-2020-judge} develops a model to fill blanks in a Mad Libs format to generate humorous sentences and \citet{yang-etal-2020-textbrewer} edit headlines to make them funny. More research is required to generate humorous sentences that are not constrained by their semantic structure.

\paragraph{Natural language explanation generation.}
Collecting and utilizing natural language explanations to help various NLP tasks is an emerging topic. The earliest work by~\newcite{ling-etal-2017-program} collected natural language justifications, called rationales, to help solve math problems. However, their setup is limited to solving math problems given how their rationales and models were structured. 
\newcite{jansen-etal-2018-worldtree} composed a dataset of explanation graphs for elementary science questions to support multi-hop inference. Like~\newcite{ling-etal-2017-program}, they emphasized the explanations structures. 
Several works have introduced large-scale datasets of natural language explanations for the natural language inference (NLI)~\cite{camburu2018snli,kumar2020nile}, commonsense reasoning~\cite{rajani2019explain}, and hate speech detection~\cite{mathew2021hatexplain} tasks. 
However, there are no existing datasets or models that focus on explaining humor, which is a challenging task that involves commonsense and world knowledge.

\paragraph{Pun detection.} 
Being able to detect puns can be an essential step to generating them. SemEval 2017 Task 7 \cite{miller-etal-2017-semeval} introduced the challenge of pun detection, location detection and sense interpretation for homographic and heterographic puns. They also released a dataset which has become the backbone of our and several other related works.~\citet{Diao2019Heterographic} make use of gated attention networks to detection heterographic puns.~\citet{zou-lu-2019-joint} introduce a tagging scheme to jointly detect and locate puns, and apply this approach to both heterographic and homographic puns. ~\citet{zhou-etal-2020-boating} jointly model contextual and phonological features into a self-attentive embedding in their approach for pun detection and location tasks.

\section{Conclusion}
In this paper, we contribute a dataset of extensive, high-quality annotations of humor explanation, keywords, and fine-grained funniness ratings. This is the first humor dataset with such extensive and fine-grained annotations. Based on the annotations, we propose two tasks: pun explanation and keyword-conditioned pun generation, to challenge state-of-the-art natural language understanding and generation models' ability to understand and generate humorous text. We benchmark several strong models' performances on the two proposed tasks to validate the practical usage of the proposed annotations, and show that our human-annotated explanations and keywords are beneficial in understanding and generating humor. Future directions include a deeper analysis of how to characterize pun explanation more objectively within our annotation scheme, as well as further exploration of better models for both the pun explanation and pun generation tasks.

\section*{Acknowledgements}
The authors would like to thank Scott Benson and the rest of the Alexa Data Services Rapid Machine Learning Prototyping (RAMP) team for all of their help with preparing and performing the annotation task. We also thank anonymous reviewers for their constructive feedback and suggestions that helped improve the paper.

\section*{Limitations}

This work focuses on understanding and generation of puns, a single and very specific form of humorous language. We hope that our annotation schema and methods can be used in the future to extend to other forms of humor, e.g., joke generation. Additionally, we acknowledge that humor is a highly subjective area, i.e., what might be perceived as humorous may differ greatly from one person to another depending on their unique backgrounds and experiences. We hope this work can be used as an initial framework to begin characterizing humor through human-written explanations, such that it can be used more broadly to give insight into what contributes to humorous content for different individuals and groups. 

Finally, since we use pretrained language models for our generation tasks, we note that this makes our models susceptible to generating biased or sensitive content. While we do not explicitly address concerns around bias/sensitive content within our framework to date, we aim to incorporate these considerations into pun generation as we develop new models, including methods to filter our inputs and generated data for toxicity and biased references that may be deemed offensive.

\section*{Ethics}

We hereby acknowledge that all of the co-authors of this work are aware of the provided \textit{ACL Code of Ethics} and honor the code of conduct.

{The text in the dataset (puns and non-pun text) is from the SemEval 2017 Task 7 dataset~\cite{miller-etal-2017-semeval} including jokes, aphorisms, and other short texts sourced from professional humorists and online collections. No user data from commercial voice assistant systems is used.} We collect the human annotation of pun keywords, explanations, and other meta-fields via full-time employees (with all employee-entitled fringe benefits) who are hired by the co-authors' organization for the purposes of data annotation and are not co-authors of the paper. We ensure that all the personal information of the workers involved (e.g., usernames, emails, urls, demographic information, etc.) is discarded in our dataset.
Overall, we ensure our pay per task is above the the annotator's local minimum wage (approximately \$15 USD / Hour).

\bibliography{anthology,hake}
\bibliographystyle{acl_natbib}

\newpage
\appendix
\section{ExPUN Dataset Annotation Details}
\label{app:annot_guidelines}

\subsection{Annotation Guidelines}
Below, we include the annotation guidelines we used to collect the ExPUN dataset. All pun texts in the provided examples are from the original SemEval 2017 Task 7 dataset~\cite{miller-etal-2017-semeval}.\footnote{\url{https://alt.qcri.org/semeval2017/task7/}. The data is released under CC BY-NC 4.0 license (\url{https://creativecommons.org/licenses/by-nc/4.0/legalcode}).}

\paragraph{Guidelines} You will be provided a CSV file of short texts, one short text to be annotated per row. Each row contains the text content as well as columns for each of the requested annotations. For each row, read the text carefully, and provide the following annotations:
\begin{enumerate}
    \item Mark  whether you understood the text with 0/1 (0 didn’t  understand, 1 understood the text).
    \begin{itemize}
        \item If you don’t  understand the meaning of the text (regardless of whether or not it  should be perceived as funny), rate the sample as 0 (didn’t understand).
        \item For this  assessment, you can use a quick Google search to look up any  vocabulary/terms you don’t immediately understand. However, if the amount  of research it would take to understand the text goes beyond a quick  (<1 min) search, rate the sample as 0 (didn’t understand).
        \item \textit{Example text  that was marked “don’t understand” (0):} A doctor's  mistakes go six feet under; a dentist's cover an acre.
        \item If you rate this  sample as 0 (didn’t understand), skip the rest of the annotation for this  sample.
    \end{itemize} 
    \item Mark  whether you find the text offensive or inappropriate with 0/1  (0 not offensive, 1 offensive), meaning the text is racist or  is biased against marginalized groups, or is generally offensive. If you rate this sample as 1 (is offensive), you may optionally skip  the rest of the annotation for this sample.
    \item Mark whether you  think the text is intended to be a joke with 0/1 (0 not a joke,  1 is a joke).
    \begin{itemize}
        \item Text should be  labeled as 1 (is a joke) even if it intends to be humorous, but falls  flat or is a lame/bad joke.
        \item \textit{Example text  labeled 0 (not a joke):} All that glistens is not gold.
        \item \textit{Example text  labeled 1 (is a joke):} These are my parents, said Einstein relatively.  \\ \textit{Why is this a joke?} Though  subtle, the text is a pun on the word “relatively” that associates  Einstein with his relatives (parents) and his theory of relativity.
        \item If you rate this  sample as 0 (not a joke), skip the rest of the annotation for this  sample.
    \end{itemize}
    \item Rate  funniness on a Likert scale of 1-5 (1 very not funny, 5 very funny).
    \begin{itemize}
        \item \textit{Score of 1:} A very not  funny joke consists of a joke that is not funny at all, or tries to  be funny but does not achieve the intended effect. \\ \textit{Example of Funniness 1 (not funny):} These are my parents, said  Einstein relatively.
        \item \textit{Score of 3:} An average  joke consists of a joke that that is average and may elicit some  chuckles (or groans) from you or others. \\ \textit{Example of Funniness 3 (average funniness):} When they told him  that his drum couldn't be fixed, it didn't resonate very well.
        \item \textit{Score of 5:} A very  funny joke consists of a good joke that you find humorous and  potentially would want to share/tell to others.\\ \textit{Example of Funniness 5 (very funny):} Yesterday I accidentally  swallowed some food coloring. The doctor says I'm OK, but I feel like  I've dyed a little inside.
    \end{itemize}
    \item Explain  in concise  natural language about why this joke is funny. If  external or commonsense knowledge is required to understand the  joke and/or its humor, please include the relevant knowledge in your  explanation. If  the joke is a pun or play on words, you must provide an  explanation of how the play on words works. 
    \begin{itemize}
        \item \textit{Example joke:} What do you use to cut a Roman Emperor’s hair?  Caesars. \\ \textit{Bad explanation:} The joke is a play on words about Caesar and  scissors. \\ \textit{Good explanation:} The joke is a play on words: Caesar was a Roman  Emperor, and ``Caesars'' sounds like ``scissors'', which is something you use  to cut hair.
        \item \textit{Example joke:} There was a kidnapping at school yesterday. Don’t  worry, though – he woke up! \\ \textit{Bad explanation:} The joke is a play on words about kidnapping →  kid napping. \\ \textit{Good explanation:} The joke is a play on words. The word  ``kidnapping'' implies that a kid was taken hostage at school, but ``he  woke up'' suggests that it was actually just a kid taking a nap  instead.
    \end{itemize}
    \item Pick  out (as few as possible) keyword phrases from the joke that are  related to the punchline/the reason of the joke being funny (written as a pipe-separated  (|) list of phrases with spaces). 
    \begin{itemize}
        \item Phrases  can be multiple words long.
        \item The  keyword phrases should be copied verbatim from the joke (no need to  reword them).
        \item Keep keyword  phrases sparse and mainly limited to content words. The  keyword phrases should not span the entire joke. As a general guideline,  the words in keyword phrases should make up <50\% of the words in the  full joke (though this may be difficult to achieve for shorter jokes).
        \item \textit{Example joke:} I used to hate maths but then I realised decimals  have a point. \\ \textit{Bad keywords (too dense):} I used to hate maths but | decimals have  a point \\ \textit{Good keywords:} maths | decimals | point
    \end{itemize}
\end{enumerate}

We note that this is a highly subjective task, since different people perceive humor differently! We encourage you to do your best to determine how to annotate each item as consistently as possible.

\paragraph{Example annotations} (funniness ratings are subjective, and may differ from yours!):
\begin{itemize}
    \item \textit{Text:} Yesterday I accidentally swallowed some food coloring. The doctor says I'm  OK, but I feel like I've dyed a little inside. \\
    \textit{Understand:} 1 \\
    \textit{Offensive:} 0 \\
    \textit{Is a Joke:} 1 \\
    \textit{Funniness:} 5 \\
    \textit{Explanation:} The joke is a pun. The main character feels they’ve  ``died a little inside'' meaning they’ve been changed for the worse by  swallowing food coloring. At the same time, food coloring contains dye, so  the main character has been ``dyed'' on the inside by swallowing some. \\
    \textit{Keywords:} swallowed | food coloring | dyed a little inside
    \item \textit{Text:} Waiter, there's  a fly in my soup! ``I know. It gives you a nice buzz doesn't it?'' \\
    \textit{Understand:} 1 \\
    \textit{Offensive:} 0 \\
    \textit{Is a Joke:} 1 \\
    \textit{Funniness:} 2 \\
    \textit{Explanation:} This is both a pun and a reference to a common joke  format. ``Waiter, there's a fly in my soup!'' is an old joke setup  with varying punchlines. Flies make a noise commonly described as a  ``buzz''. ``Buzz'' can be used as a noun referring to a  pleasant heightened sensation, commonly from drinking alcohol. \\
    \textit{Keywords:} fly | soup | buzz
    \item \textit{Text:} The evil onion  had many lairs. \\
    \textit{Understand:} 1 \\
    \textit{Offensive:} 0 \\
    \textit{Is a Joke:} 1 \\
    \textit{Funniness:} 3 \\
    \textit{Explanation:} This is a pun. An evil lair is a hideout for a villain  in a comic book or show. Onions are layered vegetables. The joke is that  the onion had many lairs because it was evil. \\
    \textit{Keywords:} evil onion | many lairs
    \item \textit{Text:} Hope for the  best, but prepare for the worst. \\
    \textit{Understand:} 1 \\
    \textit{Offensive:} 0 \\
    \textit{Is a Joke:} 0 \\
    \textit{(No need to fill in any more information in subsequent columns, as  this text is not a joke.)}
\end{itemize}

\paragraph{Additional calibrating examples} The following examples were rated with an average Funniness rating >= 2 in previous pilot rounds and can be used to calibrate your rubric for assigning Funniness scores.
\begin{itemize}
    \item \textit{Text:} Drinking too much of a certain potent potable may require a leave of  absinthe. \\
    \textit{Funniness ratings:} [3, 4, 2, 1, 2] \\
    \textit{Average rating:} 2.4
    \item \textit{Text:} Animals that tunnel in the soil have to have an escape root. \\
    \textit{Funniness ratings:} [3, 3, 1, 2, 2] \\
    \textit{Average rating:} 2.2
    \item \textit{Text:} My friend's bakery burned down last night. Now his business is toast. \\
    \textit{Funniness ratings:} [4, 3, 2, 1, 2] \\
    \textit{Average rating:} 2.4
    \item \textit{Text:} What is the best store to be in during an earthquake? A stationery store. \\
    \textit{Funniness ratings:} [1, 3, 2, 2, 3] \\
    \textit{Average rating:} 2.2
\end{itemize}

\subsection{Feedback from Pilot Rounds}
\label{app:annot_details}
{We did a few pilot rounds to help annotators calibrate on funniness, since not only is funniness highly subjective, but also since many puns aren’t ``traditionally funny'', but instead more humorous due to being ``clever'' or ``creative''. Feedback we received from annotators was mostly around including more detailed definitions and examples for highly-subjective criteria such as ``funniness'' and ``offensiveness''. We added questions on whether annotators ``understood the text'' to help distinguish between puns that were not understood vs. puns that were understood but then marked as ``not funny'', and added clarifying examples of ``joke keywords'' to discourage excessive copying of the input text in the annotations.}

\subsection{Inter-Annotator Agreement for Offensiveness ($AF_2$)}
\label{app:offensiveness}
{We note relative low inter-annotator agreement for $AF_2$, as identifying offensive/inappropriate content is a highly subjective and complex task, as it generally covers social stereotypes, biases, aggressive expressions, micro-aggressions, etc. Kappa looks at agreement of raw scores, while Spearman computes correlation of ranks. Combining these differences with the subjectivity of the task could explain the disparity between Kappa and Spearman scores for $AF_2$.}
\section{Analysis of Annotated Explanations}
\label{app:explanations}

\subsection{Frequent Explanation Keywords}

\begin{figure}[t!]
    \centering
    \includegraphics[trim=0cm 0cm 0cm 0cm, clip, width=\linewidth]{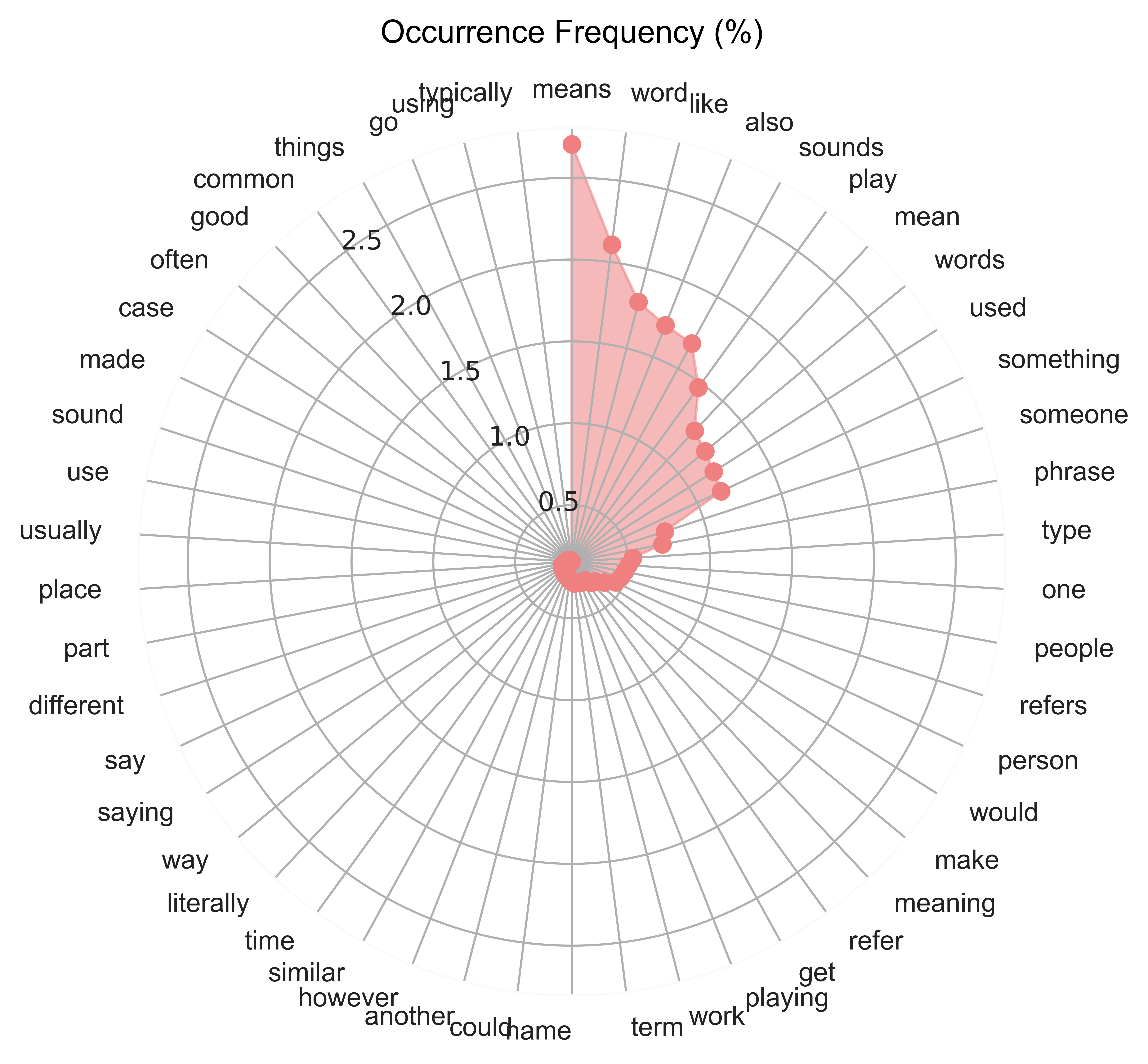}
    \caption{Top 50 most frequent words in explanations.}
    \label{fig:most_frequent}
\end{figure}

Figure \ref{fig:most_frequent} shows the distribution of the top 50 most frequent words in our annotated explanations (after removing stop words and punctuation, as well as the task-specific words ``pun'' and ``joke'').

\subsection{Explanation Sentence Templates}

To further explore what kinds of explanations annotators have provided within ExPUN, we use a simple templatization scheme to uncover common patterns. Given an input pun and explanation pair from the dataset, we templatize an explanation by replacing any content words (non-stop words) from the pun that show up in the explanation. For example, for the pun ``I wrote a novel about a fellow who had a small garden. It didn't have much of a plot.'', the explanation sentence ``This is a play on the word plot.'' would become the template ``this is a play on the word [X].'' We then count the number of unique templates across the dataset.

\begin{table}[t!]
\small
\begin{center}
\small
\begin{tabular}{ l l }
\toprule
 \textbf{Freq} & \textbf{Explanation Sentence Template} \\ \midrule
  466 & [X] sounds like [X] \\
  116 & this is a pun on the word ``[X]'' \\
  16 & [X] is a type of [X] \\
  11 & this joke is playing on the word ``[X]'' \\
  & and its different meanings \\
  8 & [X] is another word for [X] \\
  3 & [X] has two meanings \\ \midrule
  \bf Pun & \bf A gambling gardener usually hedges his bets. \\
  1 & to ``hedge your [X]'' means to take a chance on \\ 
  & something and a hedge is a type of shape that\\ 
  &  a [X] can cut plants into \\
  \bf Pun & \bf I must attend my flock, said Tom, sheepishly. \\
  1 & this is an attempt at a joke since a group of \\ 
  & sheep are known as a [X] and `sheepishly' \\
  & is another term for `embarrassed' \\
  \bf Pun & \bf People who make necklaces may get \\ 
  & beady eyes.\\
  1 & ``[X]'' here refers to the beads in [X] and \\ 
  & the way [X] have to hyper-focus their [X] when \\
  & crafting hand-made jewelry \\
\bottomrule
\end{tabular}
\end{center}
\caption{\label{table:explanation-templates} Sample explanation sentence templates collected in ExPUN, along with their frequencies.}
\end{table}

Table \ref{table:explanation-templates} shows counts frequency counts for different selected templates found in ExPUN. The top half of the table shows instances of highly-frequent explanations, such as ``[X] sounds like [X]'' indicating a heterographic pun. The bottom half of the table shows examples of unique templates that show up only once but exemplify rich explanations that include context-specific words that are useful for pun understanding. We note that the more frequent templates help to characterize common ways to explain the humor within puns, while the unique templates serve as highly-informative descriptions that can aid in the pun classification task (e.g., a detailed, contextualized definition of a word/phrase).

\section{Keyword Aggregation Algorithm}
\label{app:aggregated-keyword}
We propose the keyword aggregation algorithm in Algorithm~\ref{alg:template} to merge keywords annotation among different workers.

\begin{algorithm}[h!]
\small
\caption{Keyword Aggregation Algorithm}
\textbf{Input}: For each instance $X_i$, $i \in \{1, ..., N\}$, annotations from every worker $w_j$, $j \in \{1, ..., 5\}$ denoted as $X_{ij}$.
\\
\textbf{Output}: keywords for $X_{i}$
\begin{algorithmic}[1] 
\FOR {$j \in \{1, ..., 5\}$} 
\STATE  // calculate the reliability score \\
$S_{j}=\frac{1}{N}\sum_{i=0}^{N}{}$(\#keywords$-$\#average tokens in each keyword) 
\ENDFOR
\STATE sort all workers with $S$ and get preferred worker list $L$
\STATE // \textbf{set worker with the highest $S$ as anchor worker $w_{a}$}
\STATE aggregated\_keywords = []
\FOR  {$K_{z} \in {X_{ia}}$}
\STATE filtered\_keywords $K_{\textrm{filter}}$ = []
\FOR {$j \in \{1, ..., 5\}$}
\FOR {$K_{p} \in X_{ij}$}
\STATE calculate $F(K_{z}, K_{p})$
\ENDFOR
\STATE choose the keyword $K_{P}$ in ${X_{ij}}$ with highest $F$
\IF{$F(K_{z}, K_{P}) > 60$}
    \STATE append keyword $K_{P}$ to ${K_{\textrm{filter}}}$
\ENDIF
\ENDFOR
\STATE $AVG_{a}$ = $\frac{1}{\textrm{len}(F_{filter})}\sum F(K_{z}, K) K \in K_{\textrm{filter}}$
\STATE set the worker with the second highest $S$ as new anchor worker $w_{b}$. Repeat $L_6$-$L_{18}$ and get $AVG_{b}$
\IF{$AVG_{a} \geq AVG_{b}$}
    \STATE append $X_{ia}$ to aggregated\_keywords
\ELSE
    \STATE append $X_{ib}$ to aggregated\_keywords
\ENDIF
\STATE /* if only one worker has keyword annotation, append this worker's annotation to aggregated\_keywords */
\STATE Remove duplication from aggregated\_keywords
\ENDFOR
\end{algorithmic}
\label{alg:template}
\end{algorithm}
\section{Classifier Implementation Details}~\label{sec:appendix-classifier}
We finetune pretrained language models for classifying whether given text samples are jokes, 
and we use HuggingFace~\cite{wolf-etal-2020-transformers} throughout our implementation for accessing model checkpoints and modeling. For hyper-parameter search, we tried the combinations of learning rate \{$1e^{-4}$, $3e^{-4}$, $1e^{-5}$, $3e^{-5}$\} * training epoch \{3, 10, 20\}. The final hyperparameters for bert-base, roberta-base and deberta-base are: learning rate $1e^{-5}$, training epoch 20 and training batch size 32. For roberta-large-mnli and bart-large-mnli models, we reduce the training epochs to 3 and training batch size to 8. We choose the checkpoint with the best accuracy on the dev set for inference.

For ELV model, we use the released code  and inherited most of their default hyperparameters for \emph{ELV-sa}.\footnote{\url{https://github.com/JamesHujy/ELV/blob/main/EST_sa/train_restaurant.sh}} We change the training batch size per GPU to 4 to accelerate the training.

\section{T5 Implementation Details}
\label{sec:appendix-t5}
We finetune multiple T5 models~\cite{t5} in our work, and we use T5-base from SimpleT5~\footnote{\url{https://github.com/Shivanandroy/simpleT5}} throughout our implementation. We use 512 and 256 for the maximum source length and the maximum target length respectively.
As the optimizer, we use AdamW~\cite{adamw} with a learning rate of 0.0001. For the pretraining stage, we finetune T5 for 3 epochs on retrieved BookCorpus data. During the finetuning stage, we train each model on a Tesla V100 with a batch size of 8 for 30 epochs. 
During inference, we use beam search as the decoding method with a beam size of 2. 
We terminate decoding when the EOS token is generated or the maximum target length is reached.   

\end{document}